\documentclass[a4paper,UKenglish,cleveref, autoref, thm-restate]{lipics-v2021}
\usepackage{graphicx}
\usepackage{tikz-qtree}
\usepackage{tikz}
\usepackage{pgfplots}
\usepackage{amsmath}
\usepackage{xspace}
\usepackage{float}

\title{SATfeatPy -- A Python-based Feature Extraction System for Satisfiability} 

\author{Benjamin Provan-Bessell}
{Insight Centre for Data Science and Analytics, University College Cork, Ireland
}
{b.provanbessell@cs.ucc.ie}
{}
{}

\author{Marco Dalla}
{SFI Centre for Research Training in AI, University College Cork, Ireland
}
{m.dalla@cs.ucc.ie}
{}
{}

\author{Andrea Visentin}
{School of Computer Science \& IT, University College Cork, Ireland
}
{andrea.visentin@ucc.ie}
{}
{}

\author{Barry O'Sullivan}
{Insight Centre for Data Science and Analytics, University College Cork, Ireland
}
{b.osullivan@cs.ucc.ie}
{}
{}

\authorrunning{B. Provan-Bessell, M. Dalla, A. Visentin, and B. O'Sullivan} 

\Copyright{Benjamin Provan-Bessell, Marco Dalla, Andrea Visentin, and Barry O'Sullivan} 

\ccsdesc[100]{•Theorem proving and SAT solving\\
•Feature selection} 

\keywords{Feature extraction, Satisfiability, Python library} 

\category{} 

\relatedversion{} 

\acknowledgements{This publication has emanated from research conducted with
financial support of Science Foundation Ireland under Grant
16/RC/3918, 12/RC/2289-P2, and 18/CRT/6223, which are
co-funded under the European Regional Development Fund.
For the purpose of Open Access, the author has applied a CC BY public copyright licence to any Author Accepted Manuscript version arising from this submission.}

\EventEditors{John Q. Open and Joan R. Access}
\EventNoEds{2}
\EventLongTitle{SAT-22 (25th International Conference on Theory and Practice of Satisfiability Testing)}
\EventShortTitle{SAT-22}
\EventAcronym{SAT22}
\EventYear{2022}
\EventDate{August 2--5, 2022}
\EventLocation{Haifa, Israel}
\EventLogo{}
\SeriesVolume{25}
\ArticleNo{39}

\newcommand{\sfp}{\href{https://github.com/bprovanbessell/SATfeatPy}{SATfeatPy}\xspace}
\newcommand{\doc}{\href{https://github.com/bprovanbessell/SATfeatPy/blob/main/documentation/features.md}{documentation}\xspace}

\newcommand{\satzillab}{\textit{SATzilla base}\xspace}
\newcommand{\satzillap}{\textit{SATzilla full}\xspace}
\newcommand{\ansotegui}{\textit{ANT}\xspace}
\newcommand{\alfonso}{\textit{ALF}\xspace}
\newcommand{\all}{\textit{All}\xspace}

\begin{document}

\maketitle

\begin{abstract}
Feature extraction is a fundamental task in the application of machine learning
 methods to SAT solving. It is used in algorithm selection and configuration for solver portfolios and satisfiability classification. Many approaches have been proposed to extract meaningful attributes from CNF instances. Most of them lack a working/updated implementation, and the limited descriptions lack clarity affecting the reproducibility. Furthermore, the literature misses a comparison among the features.
This paper introduces \texttt{\sfp}, a library that offers feature extraction techniques for SAT problems in the CNF form. This package offers the implementation of all the structural and statistical features from there major papers in the field.
The library is provided in an up-to-date, easy-to-use Python package alongside a detailed
 feature description. We show the high accuracy of SAT/UNSAT and problem category classification, using five sets of features generated using our library from a dataset of 3000 SAT and UNSAT instances, over ten different classes of problems. Finally, we compare the usefulness of the features and importance for predicting a SAT instance's original structure in an ablation study. 
\end{abstract}
\section{Introduction}

Feature extraction from SAT instances remains an essential task. Features are used to train machine learning models for a variety of tasks such as algorithm selection, classification of whether an instance is satisfiable or not, and to which class an instance is drawn from. 
Most existing approaches work on instances encoded in conjunctive normal form (CNF).
While automatic feature extraction is an ongoing task, manual implementation and extraction of these features is still an effective way of judging the class of a problem. 
SATzilla~\cite{xu2008satzilla} provides the benchmark for portfolio classification of SAT problems, providing a wide range of statistical, structural and probing features of a CNF that help describe the complexity and structure of the instance. However, other features have been more recently crafted~\cite{alfonso2014new} (\alfonso), 
and~\cite{ansotegui2017structure} (\ansotegui). Feature sets across studies are not easily comparable. Furthermore, their implementations are often outdated, possibly not in a 
 executable state in their current form, and challenging to understand and use. 
\sfp, the tool we are presenting here\footnote{Please find implementation and documentation at \sfp\\ \url{https://github.com/bprovanbessell/SATfeatPy}.} re-implements these feature extractors in a single library to extract and compare them.

This paper provides a description of all features, with a more in-depth explanation available in the \doc. The code of the library is easy to read and well commented, allowing a more interested reader to investigate the implementation. Many recent studies have
focused on making SAT research and prototyping easier, boosting investigation in this area by providing functionalities in Python, e.g. PySAT~\cite{ignatiev2018pysat}. A goal of this work is to contribute to this effort.

We perform a study on five sets of features extracted with our tool from a dataset consisting of 3000 SAT and UNSAT CNFs, over ten different classes of problems. First, the feature sets are used to classify whether the problem is satisfiable (SAT) or not satisfiable (UNSAT) and to which problem class the instance belongs. The best approach reaches an accuracy of 99.8\%. An ablation study and an investigation of the most important features for classification within each set is completed. The computational time for each set of features is also analysed.

The remainder of this paper is structured as follows. First, the literature describing and motivating different sets of features is reviewed (Section \ref{sec:lit}). Section \ref{sec:feat_desecription} provides a high-level description of the features calculated in this tool. An extensive computational study presented in Section \ref{sec:experiments} compares the efficacy of different sets of features in the classification of SAT instances, evaluating their contribution and the computational effort required. Finally, Conclusions and Future work is discussed (Section \ref{sec:conclusion}).
\section{Literature Review}
\label{sec:lit}

We survey papers that devise hand-crafted features of SAT instances and utilise these features for various tasks such as algorithm selection and solver parameter optimisation. This review is split into two sections:  the first discusses the theory and motivation of hand-crafted SAT instance features, and then applications of said features are analysed.

The first paper that shows the importance of extracting structural information from SAT instances is Mitchell et al.~\cite{DBLP:conf/aaai/MitchellSL92}. They underline a  strong correlation between the hardness of a SAT instance and the ratio between the number of variables and clauses. Further work has since been done to estimate with increasing precision the empirical hardness of a SAT instance and, consequently, its likely solubility in meaningful time by a specific solver. In a later paper, 
Nudelman et al.~\cite{DBLP:conf/cp/NudelmanLHDS04}, present 84 new statistical features. These stem from nine main categories: problem size, variable-clause-graph, variable graph, clause graph, balance, proximity to Horn formulae, LP-based, DPLL-probing, and local search probing. 
Further work by Ansotegui et al.~\cite{ansotegui2017structure}
and Alfonso et al.~\cite{alfonso2014new} introduces new features to describe the structure further and help classify CNF formulas. All the features presented so far are handcrafted, a more recent branch of research focuses on automated feature extraction using 
deep learning~\cite{Dalla2021}.

Extracting representative features of a boolean formula has many practical applications. One of the most relevant is portfolio algorithm selection~\cite{portfolio}, a technique in which multiple algorithms (each of which specialises in solving/evaluating problem instances of a particular type) form a portfolio. The current problem instance is then classified, and the algorithm which performs the best on that class is then used on said problem instance. SATzilla-07~\cite{xu2008satzilla} is one of the first systems that leverages the use of these characteristics (features) to extract a measure of empirical hardness from satisfiability problems. This measure is then used to construct a per-instance algorithm portfolio that automatically selects the best SAT solver from a set of different SAT solvers, to minimise the solver's runtime on the instance. 

Further work on automated algorithm selection, as well as other related approaches like algorithm configuration, planning and portfolio selection, is presented in the survey by Kerschke et al. \cite{Kerschke2018}. Their paper also presents an overview of the different features that can guide the selection of solvers for specific instances. Xu et al. \cite{hydra2010} work combines the techniques of portfolio-based algorithm selection and automated algorithm configuration to produce a set of solvers capable of tackling instances in different problem domains. Lindauer et al. \cite{Lindauer2015} leverage on a different set of features to perform an automated algorithm configuration of a popular portfolio-based solver. Finally, Misir et al. \cite{alors2017}  build a recommender system that selects the best solver through collaborative filtering. Approaches inspired by SATzilla, where they utilise different machine learning techniques to select the optimal algorithm, can be seen in the paper from Nikolic et al. \cite{nikolic2013simple}. An example of instance classification is demonstrated in a 2008 paper by Devlin et al.\cite{Devlin2008}. They use multiple machine learning classifiers to directly predict SAT instance satisfiability by training them on the same features used by SATzilla.
Furthermore, work that leverages deep learning techniques to perform algorithm selection, feature extraction and instance classification is presented in the papers of Loreggia et al. \cite{DBLP:conf/aaai/LoreggiaMSS16}, and Bunz et al. \cite{Bunz2017}.

\section{Feature Descriptions}
\label{sec:feat_desecription}
This section will provide an overview of the features extracted by \sfp. Please refer to the \doc \footnote{\url{https://github.com/bprovanbessell/SATfeatPy/blob/main/documentation/features.md}} of the tool and the original papers~\cite{alfonso2014new, ansotegui2017structure, xu2008satzilla} for a more detailed and theoretical explanation of the features, and a guide on how to use the library.

\subsection{SATzilla features}
SATfeatPy implements the features as described in \cite{xu2008satzilla}, with some caveats. There are some minor differences between the description of the features within the paper and the implementation of said features. In these cases, the implementation has been followed, and the description has been updated accordingly. Before the feature descriptions, graphs, algorithms and statistics used to generate features will be briefly referenced and described.

The Variable Clause Graph (VCG) is a bipartite graph with a node for each variable, a node for each clause, and an edge between them whenever a variable occurs in a clause. The Variable Graph (VG) contains a node for each variable, and an edge between variables is created if they occur together in at least one clause. The bias (referred to as ratio in the \cite{xu2008satzilla}), is not a ratio, but a measure of the bias between positive and negative values. It is computed as: 
$2 \times | 0.5 - \frac{pos}{(pos + neg)}|$ where \emph{pos} refers to the number of positive occurrences, and \emph{neg} refers to the number of negative occurrences.
The Davis-Putman-Logeman-Loveland (DPLL) algorithm is often used as a basis for many SAT solvers. An implementation of this algorithm is used to generate probing features.
Statistics refer to a range of statistics including mean, coefficient of variation, minimum, maximum and entropy. Please refer to the \doc for details on which statistics are provided for which features.

SATzilla extracts features from preprocessed CNFs. This preprocessing is done by SatELite (part of Minisat) \cite{satelite}. This preprocessing generally reduces the number of variables, and can reduce or increase the number of clauses. Tautology reduction is then performed on the resulting CNF, followed by a round of unit propagation, to ensure that there are no unit clauses within the formula. Features are then extracted from this processed CNF. In some cases, this preprocessing either solves the formula, or finds it to be unsatisfiable, in which case no features are extracted from it.

\begin{itemize}
    \item Problem size features:
    \begin{itemize}
        \item Number of clauses.
        \item Number of variables.
        \item Ratio of clauses to variables.
    \end{itemize}
    \item Variable Clause Graph features:
    \begin{itemize}
        \item Variable nodes degrees statistics.
        \item Clause nodes degrees statistics.
    \end{itemize}
    \item Variable Graph features: Node degree statistics.
    \item Balance features:
    \begin{itemize}
        \item Bias of positive or negative literals in each clause statistics.
        \item Bias of positive or negative occurrences of each variable statistics.
        \item Fraction of binary and ternary clauses.
    \end{itemize}
    \item Proximity to Horn formula:
    \begin{itemize}
        \item Fraction of Horn clauses.
        \item Number of occurrences in a horn clause for each variable statistics.
    \end{itemize}
    \item DPLL Probing features:
    \begin{itemize}
        \item Number of unit propagations computed at depths 1, 4, 16, 64 and 256.
        \item Search space size estimate: mean depth to contradiction.
        \item Estimate of the log of number of nodes.
    \end{itemize}
    \item Local Search Probing Features:
    \begin{itemize}
        \item Number of steps to the best local minimum statistics.
        \item Average improvement to best in a run statistics.
        \item Fraction of improvement due to first local minimum statistics.
        \item Number of unsatisfied clauses in each local minimum mean.
    \end{itemize}
\end{itemize}

\subsection{Structure features for SAT instances classification (\ansotegui)}
\ansotegui introduces four new features, each aiming to give a meaningful representation of the structure of the problem. Three of the four features are extracted from two new graphs: the Variable Incidence Graph (VIG), and the Clause Variable Incidence Graph (CVIG). The VIG has its set of vertices the set of variables, edges and corresponding weights are assigned as follows: $w(x, y) = \sum_{x,y \in c} \frac{1}{{{|c|}\choose{2}}}$
where x and y are variable instances within the same clause.

The CVIG has its set of vertices the set of variables union the set of clauses, with edges and corresponding weights as follows: $w(x, c) = \frac{1}{|c|} \iff x \in c$ 

The four features computed by this module are as follows:
\begin{itemize}
    \item Powerlaw exponent \(\alpha\) of the number of variable occurrences. The number of occurrences of each variable (within the formula) is aggregated and formed into a sequence, which is then estimated to fit to a power-law \(f_v(k) \approx ck^{-\alpha_v} \), (based on the assumption that this sequence follows a power-law distribution.
    \item Modularity (community structure) of the best partition of the VIG. The modularity of a graph is a measure of strength of division of a network into smaller groups. The best partition (division into groups) is computed using the louvain method \cite{louvain}, and the modularity score of this partition is computed and used as a feature.
    \item Fractal dimension for the VIG and CVIG.
    The fractal dimension (self similar structure) of a graph represents the rate at which a graph will grow if it is self-similiar (a graph is self similiar if it keeps its structure after rescaling, where rescaling means replacing groups of nodes by a single node). The function \emph{N(r)} is the minimum number of circles needed of radius \emph{r} to cover the graph. This is computed by the burning by node degree method. \emph{N(r)} is calculated for \emph{r} up to a limit (16) for a graph g, and based on the assumption that $N(r) \sim r^{-d}$, the fractal dimension \emph{d} is estimated by interpolating \(log(N(r))\) vs. \(log(r)\).
\end{itemize}

\subsection{New CNF features and formula classification (\alfonso)}
\alfonso creates 8 weighted graphs from which statistics are calculated, along with an implementation of the recursive weight heuristic of variables. Theoretical descriptions provided in~\cite{alfonso2014new} were used to create all graphs.
The pseudocode present in~\cite{alfonso2014increasing} was also utilised to devise functions to create the AND, BAND, and EXO graphs. Statistics are calculated from the node degrees, and the edge weights (if present). These statistics include but are not limited to: minimum, maximum, mean, mode, median, standard deviation, number of zeros present, quartile values, rate of the mode value, and entropy. Please see the \doc for which statistics are extracted from which values.

Table \ref{tab:graphs_table} in the appendix describes the graphs created for this set of features.
\emph{V} represents the set of vertices, \emph{C} the set of clauses within the formula \emph{F}.

\begin{itemize}
    \item Statistics for each graph:
    \begin{itemize}
        \item Clause Variable (+ and -): The Clause-Variable graph(CV+ for positive literals, CV- for negative literals) connects literals with the clauses in which they appear.
        \item Variable: As in SATzilla, the Variable graph connects two variables when they appear in the same clause without considering the polarity, with weights described in Table \ref{tab:graphs_table}.
        \item Clause: As in SATzilla, the Clause graph connects clauses that share literals, with weights described in Table \ref{tab:graphs_table}.
        \item Resolution: The Resolution graph connects clauses when they produce a non-tautological resolvent.
        \item Binary Implication: A simple graph of a formula is the binary implication graph (BIG) which contains all literals as vertices, and all edges in the graph correspond to the binary clauses in the current formula.
        \item AND: The AND-gate graph represents the relation between literals that belong to an AND-gate $(l_0$ iff  $l_1 \dots l_k)$ encoded in the CNF formula.
        \item BAND: When the encoding of an AND-gate is partially present in the formula and the rest of the clauses can be added by blocked clause addition, a blocked AND-gate (BAND-gate) is recognized and the BAND-gate graph contains the relations between the literals in the recognized gate.
        \item EXO: The EXO-gate graph contains the relation between literals that belong to the exactly one literal gates encoded in the formula.
    \end{itemize}
    \item Recursive weight heuristic: This heuristic provides a score for each literal $x$ that represents the tendency whether $x$ is present in a model of the formula F. This is computed for 3 iterations.
\end{itemize}

\section{Experimental Results}
\label{sec:experiments}
This section evaluates the performances of the techniques presented herein on two different tasks: SAT/UNSAT and instance category classification. Then a comparison of the computational effort required to extract them is performed. We considered five sets of features:
\begin{itemize}
    \item \satzillab: The base features from SATzilla \cite{xu2008satzilla} (not including the probing features). 38 features.
    \item \satzillap: Base SATzilla features and probing features. 69 features.
    \item \ansotegui: Features from Ansotegui et al. \cite{ansotegui2017structure}. 4 features.
    \item \alfonso: Features from Alfonso et al. \cite{alfonso2014new}. 254 features.
    \item \all: Features from all papers combined. 327 features.
\end{itemize}

We create a dataset of 3000 CNFs equally distributed between SAT and UNSAT instances over ten problem categories for the classification tasks. The dataset is generated using CNFgen~\cite{lauria2017cnfgen}. This tool produces propositional formulas in the CNF DIMACS format that can be used as a benchmark for SAT solvers. It features several formula families (e.g. pigeonhole principle, ordering principle, k-coloring, etc.) as well as several formula transformations and the possibility of producing formulas directly from graph structures. Each set of features was used to train a Random Forest Classifier (RFC) and an XGBoost Classifier (XGB) \cite{chen2016xgboost} to predict whether the instance was SAT or UNSAT and to which category it belonged. The most relevant features (computed as the features that have the highest impurity decrease, based on the mean and standard deviation of this decrease) within the Random Forest Classifier were recorded for each set of features for both classification tasks. 

 \subsection{SAT/UNSAT Classification}
This experiment evaluates if the features extraction preserves the instance's satisfiability information. Table \ref{tab:sat/unsat_acc} shows the results of the experiment alongside the time required to extract the features. These results are the average accuracy based on ten runs of 10 cross-validation. The highest average accuracies are highlighted in bold.

\begin{table}[h]
    \caption{SAT/UNSAT classification accuracies and average computational times (in seconds) across feature sets.}
    \label{tab:sat/unsat_acc}
    \centering
    \begin{tabular}{|c|c|c|c|}
        \hline
        Feature set     & Random Forest (RFC)    & XGBoost (XGB) & Computational Time (s) \\ \hline
        \satzillab   & 91.33\%                   & 90.68\% &  \textbf{0.03} \\
        \satzillap   & \textbf{99.45}\%          & \textbf{99.54}\% & 3.18 \\
        \ansotegui      & 85.94\%                   & 81.82\% & 0.15 \\
        \alfonso        & 91.65\%                   & 93.85\%  & 119.54 \\
        \all             & 98.93\%                   & \textbf{99.54}\% & 122.89 \\
        \hline
    \end{tabular}
\end{table}

\satzillap performs the best, classifying 99.45\% and 99.54\% of the instances correctly with both the RFC and XGB, respectively. Interestingly, \all did not improve accuracy with the RFC but lowered it by 0.52\% compared to the \satzillap feature set. This is most likely due to the decision tree branching on different features, which provides slightly worse decisions and lower classification accuracy.
For XGBoost classification, the \all set has the same accuracy of \satzillap. These feature sets significantly outperform the other feature sets by a minimum of 5.69\%. 
Regarding the computational times, \satzillab is considerably faster than all the other feature sets. The probing process slows \satzillap considerably, and the complex graph creation makes \alfonso the most computationally expensive.
Overall, this experiment shows that the features selected by Xu et al. \cite{xu2008satzilla} extract relevant information on the instance's satisfiability. The contribution of the other features does not seem to be relevant.

We retrained the RFC over the \all dataset with 10 000 estimators to evaluate the features' importance. Surprisingly, they are all graph statistics from the \alfonso feature set, which performances are inferior to \satzillap. However, 5 of the top 10 features are taken from the Variable Clauses Graph from \alfonso, which is a modified version of the Variable Clauses Graph that is part of \satzillab. See Table \ref{tab:feat_importance_all_cat} in the appendix for more detailed results on the top 10 features.


\subsection{Problem Category Classification}
Understanding the instance's type is fundamental in algorithm portfolio approaches. The results for this multi-class classification task are available in Table \ref{tab:cat_acc}. 

\begin{table}[h]
    \caption{Problem Category Classification accuracies across feature sets, for both Random Forest and XGBoost classifiers.}
    \label{tab:cat_acc}
    \centering
    \begin{tabular}{|c|c|c|}
        \hline
        Feature set     & Random Forest (RFC)        & XGBoost (XGB) \\ \hline
        \satzillab   & 99.65\%                   & 99.66\% \\
        \satzillap   & 99.59\%                   & 99.74\% \\
        \ansotegui      & 93.14\%                   & 93.47\% \\
        \alfonso        & 99.79\%                   & 99.64\% \\
        \all             & \textbf{99.83}\%          & \textbf{99.81}\% \\
        \hline
    \end{tabular}
\end{table}

The \all set performed the best with both classifiers, with 99.83\% and 99.81\% accuracy for both RFC and XGB, respectively. This suggests that the feature sets help classify different types of instances correctly, contributing to overall better results. The task has more target labels compared to the previous one; however, all the approaches reach higher accuracy. \satzillab, \satzillap,  and \alfonso perform excellently, showing that statistical features represent the problem structure more than its satisfiability. 

The RFC was retrained over the \all dataset with 10 000 estimators to evaluate the features' importance. Interestingly, the majority of the features are from the \alfonso set, with exception of the most important feature being a \satzillap probing feature. See Table \ref{tab:feat_importance_all_cat_inst} in the appendix for detailed results on the top 10 features.


\subsection{Feature computation time}
\begin{figure}
    \centering
    \includegraphics[width=0.8\columnwidth]{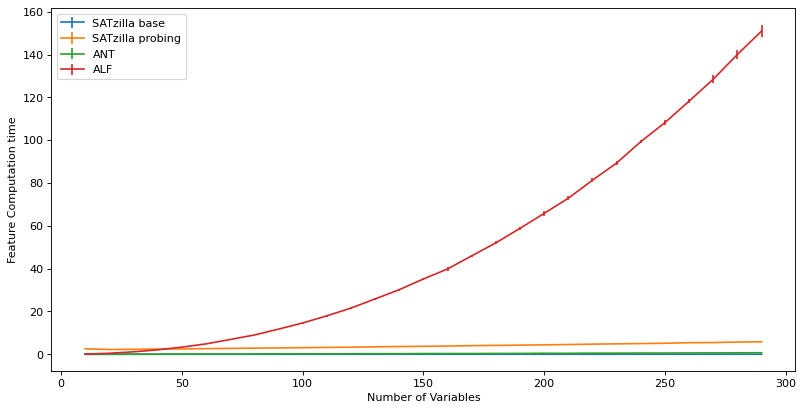}
    \caption{Time to generate features vs. size. The mean time to solve each set of instances is plotted in a line, and the variance is shown in the error bars.}
    \label{fig:time_vs_size}
\end{figure}

This experiment evaluates the scalability of the techniques. We record the computational time to generate the features over a dataset of randomly-generated CNFs. The dataset comprises 300 instances with consistent clauses to variables ratio of 4.2 and uniformly increased size (where size is the number of variables). The time to compute each feature set is plotted in Figure \ref{fig:time_vs_size}.

The effort required to compute \alfonso features increases polynomially with the problem size, making it unsuitable for big instances. This is due to the many graphs that have to be created and the features number. The other approaches increase linearly with the instance size, with \satzillap that has a minimum computational time represented by the probing timeout. \satzillab is considerably faster than the other approaches.
The appendix contains a plot without \alfonso (Figure \ref{fig:time_vs_size_no_alf}) that shows that better displays the differences between the other approaches.

\section{Conclusions and Future Work}
\label{sec:conclusion}
This paper presents (\sfp), a Python tool that re-implements CNF statistical feature extraction techniques available in the literature. The features are available in an accessible and easy-to-use library; each feature is provided with a readable implementation alongside clear documentation. An ablation study showed high classification accuracy achievable with the features extracted by this tool and the importance of each set of features. A time analysis showed the time needed to compute each feature set as problem size grew.


The tool aims to streamline the utilisation of existing SAT features and boost the development and comparison of new ones. It can be used for fast prototyping and ideas validation of portfolio solutions. The readable and updated code and the documentation clarify existing inconsistencies. 

Our team commits to keeping the code updated and adding new features introduced in the literature. We plan to do a deeper analysis of existing features in future works. For example, we plan to select subsets of features particularly effective for the specific classification task. Furthermore, we want to extend the library with different preprocessing approaches that can be used in combination with features extraction techniques. 



\bibliography{ref.bib}

\begin{thebibliography}{10}

\bibitem{alfonso2014increasing}
Enrique~Matos Alfonso.
\newblock Increasing the robustness of sat solving with machine learning
  techniques.
\newblock 2014.

\bibitem{alfonso2014new}
Enrique~Matos Alfonso and Norbert Manthey.
\newblock New cnf features and formula classification.
\newblock In {\em POS@ SAT}, pages 57--71, 2014.

\bibitem{ansotegui2017structure}
Carlos Ans{\'o}tegui, Maria~Luisa Bonet, Jes{\'u}s Gir{\'a}ldez-Cru, and Jordi
  Levy.
\newblock Structure features for sat instances classification.
\newblock {\em Journal of Applied Logic}, 23:27--39, 2017.

\bibitem{louvain}
Vincent~D Blondel, Jean-Loup Guillaume, Renaud Lambiotte, and Etienne Lefebvre.
\newblock Fast unfolding of communities in large networks.
\newblock {\em Journal of Statistical Mechanics: Theory and Experiment},
  2008(10):P10008, Oct 2008.

\bibitem{Bunz2017}
Benedikt Bünz and Matthew Lamm.
\newblock Graph neural networks and boolean satisfiability.
\newblock {\em arXiv}, pages 1--9, 2017.

\bibitem{chen2016xgboost}
Tianqi Chen and Carlos Guestrin.
\newblock Xgboost: A scalable tree boosting system.
\newblock In {\em Proceedings of the 22nd acm sigkdd international conference
  on knowledge discovery and data mining}, pages 785--794, 2016.

\bibitem{Dalla2021}
Marco Dalla, Andrea Visentin, and Barry O'Sullivan.
\newblock Automated sat problem feature extraction using convolutional
  autoencoders.
\newblock pages 232--239. Institute of Electrical and Electronics Engineers
  (IEEE), 12 2021.

\bibitem{Devlin2008}
David Devlin and Barry O'Sullivan.
\newblock Satisfiability as a classification problem.
\newblock {\em Proc. of the 19th Irish Conf. on Artificial Intelligence and
  Cognitive Science}, 2008.

\bibitem{satelite}
Niklas E{\'e}n and Armin Biere.
\newblock Effective preprocessing in sat through variable and clause
  elimination.
\newblock In {\em International conference on theory and applications of
  satisfiability testing}, pages 61--75. Springer, 2005.

\bibitem{ignatiev2018pysat}
Alexey Ignatiev, Antonio Morgado, and Joao Marques-Silva.
\newblock Pysat: A python toolkit for prototyping with sat oracles.
\newblock In {\em International Conference on Theory and Applications of
  Satisfiability Testing}, pages 428--437. Springer, 2018.

\bibitem{Kerschke2018}
Pascal Kerschke, Holger~H. Hoos, Frank Neumann, and Heike Trautmann.
\newblock Automated algorithm selection: Survey and perspectives.
\newblock {\em Evolutionary Computation}, 27:3--45, 2018.

\bibitem{lauria2017cnfgen}
Massimo Lauria, Jan Elffers, Jakob Nordstr{\"o}m, and Marc Vinyals.
\newblock Cnfgen: A generator of crafted benchmarks.
\newblock In {\em Proceedings of {SAT}}, pages 464--473, 2017.

\bibitem{portfolio}
Kevin Leyton-Brown, Eugene Nudelman, Galen Andrew, Jim McFadden, Yoav Shoham,
  et~al.
\newblock A portfolio approach to algorithm selection.
\newblock In {\em IJCAI}, volume~3, pages 1542--1543, 2003.

\bibitem{Lindauer2015}
Marius Lindauer, Holger~H Hoos, Frank Hutter, and Torsten Schaub.
\newblock Autofolio: An automatically configured algorithm selector, 2015.

\bibitem{DBLP:conf/aaai/LoreggiaMSS16}
Andrea Loreggia, Yuri Malitsky, Horst Samulowitz, and Vijay~A. Saraswat.
\newblock Deep learning for algorithm portfolios.
\newblock In {\em Proceedings of {AAAI}}, pages 1280--1286, 2016.

\bibitem{DBLP:conf/aaai/MitchellSL92}
David~G. Mitchell, Bart Selman, and Hector~J. Levesque.
\newblock Hard and easy distributions of {SAT} problems.
\newblock In {\em Proc. of {AAAI}}, pages 459--465, 1992.

\bibitem{alors2017}
Mustafa Mısır and Michèle Sebag.
\newblock Alors: An algorithm recommender system.
\newblock {\em Artificial Intelligence}, 244:291--314, 3 2017.

\bibitem{nikolic2013simple}
Mladen Nikoli{\'c}, Filip Mari{\'c}, and Predrag Jani{\v{c}}i{\'c}.
\newblock Simple algorithm portfolio for sat.
\newblock {\em Artificial Intelligence Review}, 40(4):457--465, 2013.

\bibitem{DBLP:conf/cp/NudelmanLHDS04}
Eugene Nudelman, Kevin Leyton{-}Brown, Holger~H. Hoos, Alex Devkar, and Yoav
  Shoham.
\newblock Understanding random {SAT:} beyond the clauses-to-variables ratio.
\newblock In Mark Wallace, editor, {\em Proceedings of {CP}}, pages 438--452,
  2004.

\bibitem{hydra2010}
Lin Xu, Holger~H Hoos, and Kevin Leyton-Brown.
\newblock Hydra: Automatically configuring algorithms for portfolio-based
  selection.

\bibitem{xu2008satzilla}
Lin Xu, Frank Hutter, Holger~H Hoos, and Kevin Leyton-Brown.
\newblock Satzilla: portfolio-based algorithm selection for sat.
\newblock {\em Journal of artificial intelligence research}, 32:565--606, 2008.

\end{thebibliography}
\clearpage
\appendix
\section{Graph structures for Alfonso et al.}
\begin{table}[h]
    \caption{Graph structure for graphs created in \cite{alfonso2014new}. C represents the set of clauses and V the set of variables within the formula F. Please refer to pseudocode in \cite{alfonso2014new} for an explanation of how the naive constraint graphs (AND, BAND, EXO) are constructed.}
    \label{tab:graphs_table}
    \centering
    \begin{tabular}{|c|c|c|c|}
        \hline
        Graphs      & Vertices  & Edges                     & Weights \\ \hline
        $CV^+, CV^-$  & $V \cap C$  & $\langle i,j \rangle$ and $\langle j,i \rangle$ iff $|j| \in C_i$ & - \\ 
        Variables   & V         & $\langle i,j \rangle$ iff there is a $1 \leq k \leq |V|, \{i,j\} \subseteq (C_k \cup C_k) $ & $2^{-k}$ \\ 
        Clauses     & C         &  $\langle i,j \rangle$ iff $C_i \cap C_j \neq \emptyset$ & $|C_i \cap C_j|$ \\
        Resolution  & C         & $\langle i,j \rangle$ iff $|C_i \cap C_j| = 1$ & $2^{-|C_i \cup C_j| -2}$ \\
        BIG         & $V(\pm)$  & $\langle \bar{i},j \rangle$ and $\langle \bar{j},i \rangle$ iff $\{i,j\} \in F$ & $2^{-|C_i \cup C_j| -2}$\\
        AND         & $V(\pm)$ & for each $l_0 \iff l_1 \land \dots \land l_k$ we add all edges $\langle l_0,l_i \rangle $ & $2^{-k}$\\ 
        BAND        & $V(\pm)$  & for each $l_0 \Rightarrow l_1 \land \dots \land l_k$ we add all edges $\langle l_0,l_i \rangle$ & $2^{-k}$\\ 
        EXO         & $V(\pm)$  & for each $EXO(l_1 \dots l_k)$ we add for each $i \neq j$ add edges $\langle l_i,l_j \rangle$ & -\\ \hline
        
    \end{tabular}
\end{table}
\section{Feature Importance}
\begin{table}[H]
    \caption{The ten most important features in Random Forest Classifier trained with 10000 estimators on SAT/UNSAT Classification.}
    \label{tab:feat_importance_all_cat}
    \centering
    \begin{tabular}{|c|c|}
    \hline
        Feature name        & Feature Importance SAT/UNSAT \\ \hline
        VCG - weights mode (\alfonso) & 0.1092\\
        BAND weights zeros (\alfonso) & 0.1021\\
        VCG + weights q1 (\alfonso) & 0.0958\\
        VCG + weights q2 (\alfonso) & 0.0935\\
        RG node zeros (\alfonso) & 0.0879\\
        VCG - weights minimum (\alfonso) & 0.394\\
        EXO weights value rate (\alfonso) & 0.0118\\
        VCG + weights minimum (\alfonso) & 0.0107\\
        BAND node minimum (\alfonso) & 0.0088\\
        CG weights q1 (\alfonso) & 0.0079\\
        \hline
    \end{tabular}
\end{table}

\begin{table}[H]
    \caption{The ten most important features in RFC trained with 10000 estimators on Category Classification.}
    \label{tab:feat_importance_all_cat_inst}
    \centering
    \begin{tabular}{|c|c|}
    \hline
        Feature name        & Feature Importance Categories \\ \hline
        SAPS First Local  (\satzillap) & 0.0241\\
        EXO node minumum (\alfonso) & 0.0203\\
        BAND node q2 (\alfonso) & 0.0196\\
        RG node value rate (\alfonso) & 0.0184\\
        BIG node value rate (\alfonso) & 0.0181\\
        VCG + weights q3 (\alfonso) & 0.0179\\
        VCG + weights zeros (\alfonso) & 0.0170\\
        CG weights q1 (\alfonso) & 0.0167\\
        BAND node max (\alfonso) & 0.0160\\
        VG node value rate (\alfonso) & 0.0159\\
        \hline
    \end{tabular}
\end{table}
\section{Feature Computation Time - No \textit{ALF} Features}
\begin{figure}[h]
    \centering
    \includegraphics[width=0.99\columnwidth]{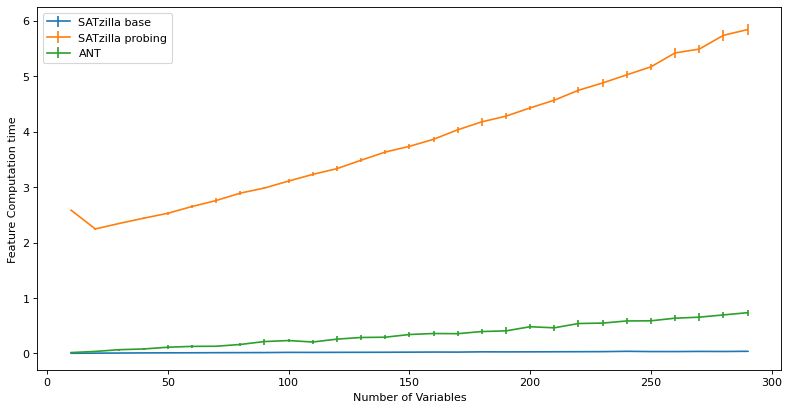}
    \caption{Time to generate features vs. size. The mean time to solve each set of instances is plotted in a line, and the variance is shown in the error bars. \textit{ALF} is excluded to display the trend of the other curves.}
    \label{fig:time_vs_size_no_alf}
\end{figure}

\end{document}